\documentclass[notitlepage,12pt]{article}

\usepackage{amsmath}

\usepackage{amssymb}
\usepackage[dvips]{graphics}
\usepackage{apalike}

\begin{document}




\title{
Concerning the Differentiability \\of the Energy Function
\\ in Vector Quantization Algorithms}

\author{\small Dominique Lepetz (1)\\ \small Max N\'{e}moz-Gaillard (2)\\ \small Micha\"{e}l Aupetit (3)\\ \\ 
\small (1) EMA-DM, 6 Av. de Clavi\`{e}res,30319 Al\`{e}s cedex, FRANCE \\ \small dlepetz at wanadoo dot fr\\ \\
\small (2) EMA-CMGD, 6 Av. de Clavi\`{e}res, 30319 Al\`{e}s cedex, FRANCE \\ \small Max.Nemoz-Gaillard at ema dot fr\\ \\
\small (3) Corresp. author - CEA, DASE, BP.12, 91680 Bruy\`eres-le-Ch\^atel, FRANCE \\ \small michael.aupetit at cea dot fr}

\maketitle

\newpage

\begin{abstract}
 The adaptation rule for Vector Quantization algorithms,
and consequently the convergence of the generated sequence,
depends on the  existence and properties of a function called the
energy function, defined on a topological manifold. Our aim is to
investigate the conditions of existence of such a function for a
class of algorithms examplified by the initial "K-means"
\cite{MacQueen67} and Kohonen algorithms
\cite{Kohonen82,Kohonen88}. The results presented here supplement
previous studies, including \cite{Tolat90}, \cite{Erwin92},
\cite{Cott94},\cite{Pages93} and \cite{Cott98}. Our work shows that the energy
function is not always a potential but at least the uniform limit
of a series of potential functions which we call a
pseudo-potential. It also shows that a large number of existing
vector quantization algorithms developed by the Artificial Neural
Networks community fall into this category. The framework we
define opens the way to study the convergence of all the
corresponding adaptation rules at once, and a theorem gives 
promising insights in that direction. We also demonstrate that the "K-means"
energy function is a pseudo-potential but not a potential in general. 
Consequently, the energy function associated to the "Neural-Gas" 
is not a potential in general.
\end{abstract}


\section*{Keywords}
Vector Quantization, K-means, Self-Organizing Maps, Neural-Gas,
energy function, potential function, pseudo-potential


\section{Introduction}
In vector quantization theory \cite{Gray98}, a set of prototypes
\footnote{also called "codebook vectors", "reference vectors", "units" or "neurons".} 
$\underline{w}=(w_{1},...,w_{n})$ is placed on a manifold
$V\subset \mathbb{R}^d, d\geq1$, in order to minimize the
following integral function, called the "energy function":

\begin{equation}\label{1}
 E_V(\underline{w})=\int_{V}\frac{1}{2}\sum_{p=1}^{n} P(v)\psi_{p}(\underline{w},v)(v-w_{p})^2dv=\int_{V}F(\underline{w},v)dv
\end{equation}
where $P(v)$ indicates the probability density defined on $V$. We
focus on the stochastic iterative approaches where at each time
step, a datum $v$ is drawn from the probability density function (pdf) $P$, and
the prototypes $\underline{w}$ are adapted
 according to $v$ using the adaptation rule:
\begin{equation}\label{adaptationrule}
\Delta w_{p}=\alpha\psi_{p}(\underline{w},v)(v-w_{p})
\end{equation}
where the adaptation step is tuned using the parameter $\alpha$
generally decreasing over the time ($\alpha$ is taken thereafter
equal to 1 without restricting the general results), and
$\psi_{p}$ is a "neighborhood" function particular to each vector
quantization algorithm. Here we focus on discontinuous $\psi_{p}$
functions.

A main concern in the field of Vector Quantization, is to decide
whether the adaptation rule (\ref{adaptationrule}) corresponds or
not to a stochastic gradient descent along the energy function
(\ref{1}), \textit{i.e.} whether this energy function is or is not
a potential onto the entire manifold $V$. On one hand, if the
energy function is a potential then the convergence of the
prototypes obeying their adaptation rule toward a minimum
of this energy function is well established, in particular in the
stochastic optimization framework \cite{Robbins51,Albert67} with
which this paper is concerned. For example, the energy function associated to
the K-means algorithm \cite{MacQueen67,Ahalt90}, stochastic version of
the LBG algorithm of Linde {\it et al.} \cite{Linde80}, is a potential as long as the
pdf $P$ is continuous \cite{Kohonen91,Pages93,Cott98}. 

On the other hand, if the energy function is not a potential, then
very few is known about the convergence of the corresponding
adaptation rule. For example, several results \cite{Tolat90,Erwin92,Heskes93,Heskes99} have already shown
that for a continuous density $P$, the corresponding vector
adaptation rule of the Kohonen Self-Organizing Map (SOM) algorithm \cite{Kohonen82}
\cite{Kohonen88} does not correspond to a stochastic gradient
descent along a global energy function, and the convergence,
although being observed in practice, turns out to be very
difficult to prove, not to mention that most of the efforts have
been carried out on the Kohonen rule \cite{Cott94,Cott98,Benaim98}.

All the vector quantization
algorithms we study in this paper 
are variants of the K-means algorithm as we will see
in section \ref{sectExistingRule}. 
We know these algorithms converge in practice toward
acceptable value of their energy functions whenever they are proved to be
associated or not to potentials.
However, the theoretical study of their convergence is not available, so they
remain largely heuristics.
Among all these algorithms, the Neural-Gas \cite{Martinetz94} deserves a particular attention.
It has been claimed by its authors 
to be associated to a global potential in general, hence to a converging adaptation rule. 
We propose a counter-example with a discontinuous pdf $P$ which demonstrates that this claim is not true. 
This shows that the study of the convergence of all these algorithms
is still in its infancy and motivates the present work.

In this paper, we propose a framework which encompasses all these algorithms.
We study this framework and we
demonstrate that the energy function associated to these
algorithms is not a potential in general.
We also demonstrate that this energy function
belongs to a broad class of functions which includes
potential functions as a special case. The energy
functions within this class are called "pseudo-potentials".
 The results we obtain do not depend on the continuity of the
probability density function $P$, and give a first step toward an
explanation why all the algorithms shown to belong to this framework
succeed, in practice, in minimizing their associated energy
function whether they are potentials or not. This framework should
open up further avenues for a general study of the convergence
properties of all the algorithms it contains at once.

In section 2, we present the framework of this study. In section
3, we define  a "pseudo-potential" function, which can be
approximated by a series of potential functions: we define the
concept of cellular manifold and this series of potentials. In
section 4, we give the main theorem which states that an energy
function of that framework is necessarily a pseudo-potential. 
We consider the K-means to show that pseudo-potentials are not always potentials.
We discuss the consequence on the convergence
of the corresponding adaptation rule.  In section 5, we show that most
of the common vector quantization algorithms
 belong to that framework. At last we
conclude in section 6.

\section{Framework}

We consider $(\mathbb{R}^{d},\parallel . \parallel)$ is the
euclidean $d$-dimensional space associated to the euclidean norm.
Let \emph{D} be a non-empty bounded set in $\mathbb{R}^{d}$. Let
$\delta$ be the diameter of \emph{D} and $V$ a topological
manifold included in \emph{D}. Let
$\underline{w}=(w_{1},...,w_{n})$ be a set of prototypes in
\emph{D}.

For p=1,...,n, the  Vorono\"{\i} cell associated to $w_p$ is
usually defined as \cite{Okabe92}:
\begin{equation}\label{2}
 V_p=\{v \in V \mid \forall \; q=1,...,n\;
 \parallel v-w_p\parallel \; \leq \; \parallel v-w_q\parallel\}
\end{equation}
The set of $V_p$\,, p=1,...,n provides a cellular decomposition of
$V$.

For any $l$, the distance between $w_l$ and $v$ is denoted
$d_{l}=\|w_l-v\|$.

 We will show in section
\ref{sectExistingRule} that the neighborhood function $\psi_p$ of
various algorithms is constructed on the basis of the Heaviside
step function of the distances $d_{l}$, denoted $H$ such that
$H(x<0)=0$ and $H(x\geq0)=1$. These step functions cause
discontinuities of the corresponding energy functions or their
derivatives, which appear at the Vorono\"\i\ cells boundaries.
This is the reason why we focused on the following class of
neighborhood functions
 in the definition of our framework:

\begin{equation}\label{functionpsi}
 \psi_{p}(\underline{w},v)=\phi_{p}(\{H(d_{l}^2-d_{m}^2)\}_{lm})
\end{equation}
where $\phi_{p}$ is a bounded function.

We consider any probability density function $P$ such that:
$\int_V P(v) dv=1$. In other words, all the results presented hereafter do not depend on the continuity of $P$.

\section{Cellular manifolds and pseudo-potential}

The discontinuities of the neighborhood functions
$\psi_{p}(\underline{w},v)$ occur onto the boundaries of the
Vorono\"\i\ cells. We shall consider a part of the manifold $V$
called \textsl{cellular manifold} (and its complementary part
called \textsl{tubular manifold}) which does not contain these
boundaries to isolate them and to ease their study. This leads to
the subsequent definition of pseudo-potential functions.

\subsection{The family of cellular manifolds $V^{\eta}$}
The cellular manifold is based on the Vorono\"{\i} cells defined
by the set of vectors $w_p$ and which is arbitrarily close to the
manifold $V$ in the sense of the Lebesgue measure.

Let $\eta$ be a number $\ll$ 1 ; we denote
$T_p^{\eta}(\underline{w})=T_p^{\eta}$ the open tubular
neighborhood, of thickness $\eta$, of the boundary of $V_p$,
included in $V$.

\begin{figure}
  \center
  \resizebox{3.5in}{!}{\includegraphics{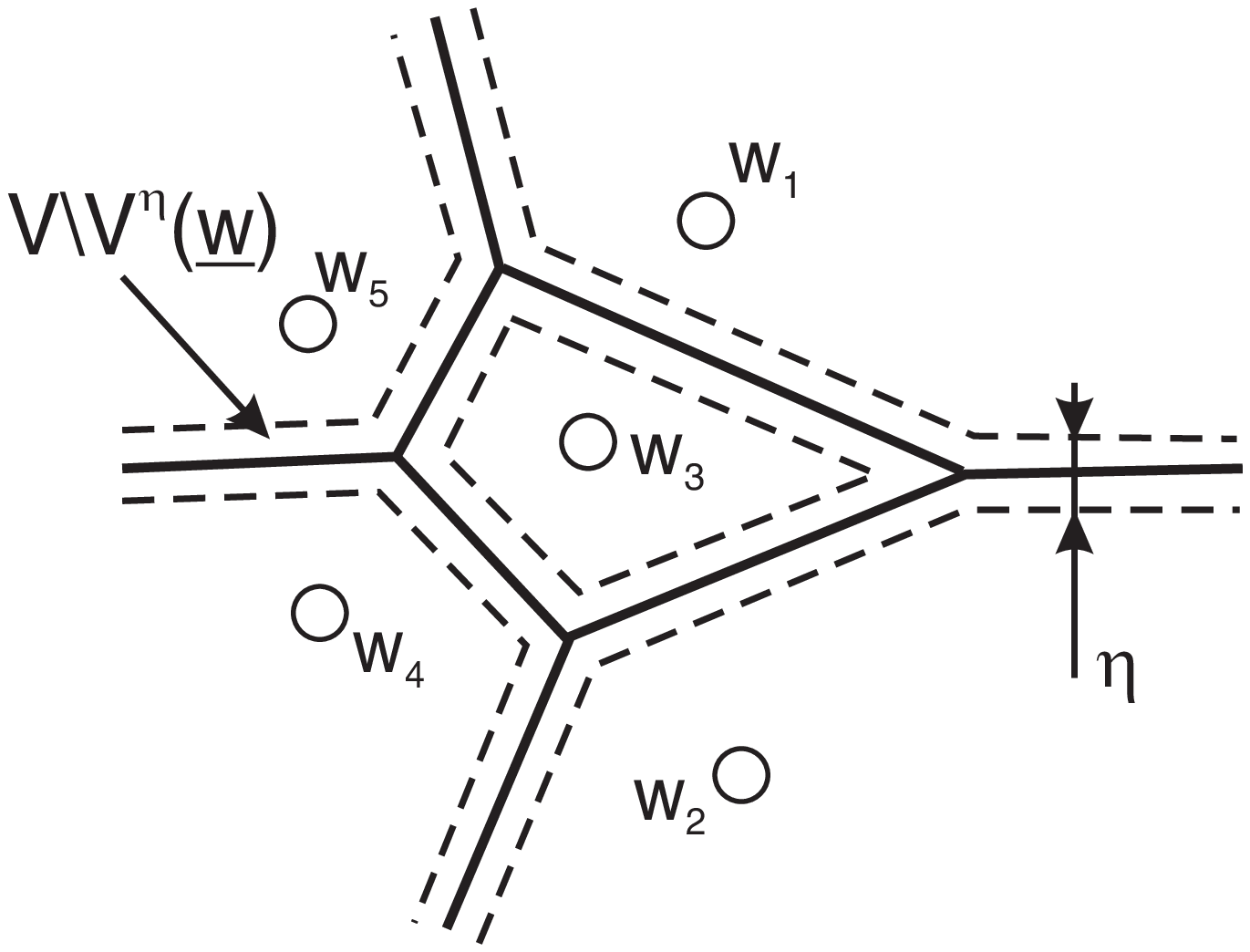}}  
  \caption{\label{figCellTubManifold}\textbf{Cellular and tubular
  manifolds}. The circles are the prototypes $\underline{w}$.
  We define tubular manifolds $V\setminus V^{\eta}(\underline{w})$
  of thickness $\eta$ (dotted lines) which contain the boundaries of
  the Vorono\"\i\ cells (plain lines), and cellular manifolds
  denoted $V^{\eta}(\underline{w})$ complementary to the tubular
  manifolds.}
  \end{figure}

This neighborhood is shown on figure \ref{figCellTubManifold} in
$\mathbb{R}^2$.

Then for a given $\underline{w}=(w_1,...,w_n)$ with $w_i \in V$,
we define the cellular manifold $V^{\eta}(\underline{w})$ as the
set of vectors of $V$ which are not in the tubular neighborhood
$T_p^{\eta}$ for all $p$:

\begin{equation}\label{definitionVeta}
 V^{\eta}(\underline{w})= V\setminus(\bigcup_{p=1}^n T_p^{\eta})
\end{equation}

That means the smaller $\eta$, the closer to $V$ the cellular
manifold $V^{\eta}(\underline{w})$ which does not contain the
boundaries of the Vorono\"\i\ cells. In other words, $V\setminus
V^{\eta}(\underline{w})$ "tends" towards the boundaries of the
Vorono\"\i\ cells while $\eta$ tends towards $0$. $V\setminus
V^{\eta}(\underline{w})$ is called \textsl{tubular manifold}.

We can then state the following property:

$\mathbb{R}^d$ being provided with the product-measure of
Lebesgue, $V^{\eta}(\underline{w})$ verifies:
\begin{equation}\label{mesureenodeeta}
  meas(V \setminus V^{\eta}(\underline{w}))=\emph{O}(\eta)
\end{equation}

And we have in particular:
\begin{equation*}
  \lim_{\eta\rightarrow 0}(meas(V \setminus V^{\eta}(\underline{w})))=0
\end{equation*}

The proof of this property follows:

\begin{align}
meas(V \setminus V^{\eta}(\underline{w}))=meas(\bigcup_{p=1}^n
(T_p^{\eta})) \leq \sum_{p=1}^n meas(T_p^{\eta})\notag
\end{align}

We have to consider two subcases according to the dimension $d$:

If d=1 : the boundaries of Vorono\"{\i} cells are points thus
their measure is null. One has in this case: $\sum_{p=1}^n
meas(T_p^\eta) \leq (n-1)\eta$, whence the result.

 If d$>$1 : we have
$meas(T_p^{\eta})\leq\,
\eta\,.\,meas_{\mathbb{R}^{d-1}}(boundary(V_p)) +
\emph{O}(\eta^2)$ where the residual term $\emph{O}(\eta^2)$ is
bounded by the following sum: each term of the sum is the product
of the measures of the (d-k) cells ($k>1$) of the polyhedral
decomposition of the boundary of $V_p$ by the volumes of the
k-balls of radius $\eta$ (i.e.
$\frac{\pi^{\frac{k}{2}}\eta^k}{\Gamma(\frac{k}{2}+1)}$). However
\textsl{D} is bounded, therefore all the measures of the
boundaries of $V_p$ are finished, whence the result.

\subsection{Definition of a pseudo-potential}

In general, a potential is defined as a differentiable function of
its variables. We define a wider class of functions that we call
{\em pseudo-potentials}, which contains potential functions as a
special case. Pseudo-potentials do not verify in general the
hypotheses of differentiability at every point but may be
approached by a series of potential functions. Thus a potential is
a pseudo-potential but the converse is false: a pseudo-potential
is not necessarily differentiable everywhere and therefore is not
necessarily a potential.

\textbf{Definition:} let $\Omega$ be a non empty and bounded set
in $\mathbb{R}^d$ and $n\geq 1$ fixed. The function $E_{\Omega}$:
$\Omega^n \rightarrow \mathbb{R}$ is called
\textsl{pseudo-potential} if there exists a family of potential
functions $E^{\eta}_{\Omega}$: $\Omega^n \rightarrow
\mathbb{R},\eta>0$, such that
\begin{equation*}\
  \lim_{\eta\rightarrow 0}\parallel
  E_{\Omega}-E^{\eta}_{\Omega}\parallel_{\infty}=0
\end{equation*}
where $\parallel.\parallel_{\infty}$ denotes the norm of the
uniform convergence.

In our case, we focus on the energy function
$E_{\Omega}(\underline{w})$ defined by (\ref{1}).

 Introducing
pseudo-potentials enables all these algorithms to be placed in the
same framework (see section \ref{sectExistingRule}). In this framework,
the neighborhood function belongs to the family defined in
(\ref{functionpsi}) and the associated energy function may not
be differentiable on the boundaries of the Vorono\"\i\ cells,
hence is possibly not a potential on the whole manifold $V$.

Which leads us to the main result about the energy function $E_V$.

\section{The energy function $E_V$ is a pseudo-potential}
\label{SectTheorem}

We show that the energy function $E_V$ defined in (\ref{1}) under
the hypotheses of the section 2, may be considered as the limit of
a series of differentiable functions over the manifold $V$,
without being itself differentiable over $V$, {\it i.e.} $E_V$ is
a pseudo-potential.

\begin{tabular}{ll}
\textbf{Theorem:} & \hspace{15mm} The energy function $E_V$ is a
pseudo-potential \\
 & \hspace{15mm} with  $\parallel
E_{V}(\underline{w})-E^{\eta}_{V}(\underline{w})\parallel_{\infty}=\emph{O}\:(\eta)$.
\end{tabular}

The first part of the theorem means $E_V$ is not necessarily a
potential over $V$, being not always differentiable on the
boundaries of the Vorono\"\i\ cells.

The second part means that the difference between the energy
 $E_V(\underline{w})$ and the energy
 $E_V^{\eta}(\underline{w})$
 both defined on the whole $V$, is bounded by a
 value proportional to $\eta$, hence as small as wanted.
 In other
 words, even if $E_V(\underline{w})$ is not a potential, it
 is very close to be one.

\subsection{Proof of the theorem}

To prove that $E_V$ is a pseudo-potential, we consider for
$\eta>0$, the functions $E^{\eta}_V$ defined as:
\begin{equation*}
\forall \underline{w} \in V^n,\;
E^{\eta}_V:\underline{w}\longrightarrow\int_{V^{\eta}(\underline{w})}F(\underline{w},v)dv
\end{equation*}

Then we first show that these functions which are defined on the
whole manifold $V$, are differentiable on $V^\eta(\underline{w})$
({\it i.e.} the domain where the integral is carried out). And
second, we show that the difference $\parallel
E_{V}(\underline{w})-E^{\eta}_{V}(\underline{w})\parallel_{\infty}$
equals $\emph{O}\:(\eta)$, hence that $\lim_{\eta\rightarrow
0}\parallel E_{\Omega}-E^{\eta}_{\Omega}\parallel_{\infty}=0$,
fulfilling the conditions necessary for $E_V$ to be a
pseudo-potential.

The proof of the first part of the theorem rests on the behavior
of the functions $\psi_p$. When the current $v$ are far enough
from the boundaries of the Vorono\"{\i} cells, these functions
behave like constants, while onto these boundaries they have
discontinuities. We need to insure the differentiability according
to $\underline{w}$ of the $F(\underline{w},v)$ functions which
depend on the $\psi_p$ functions, and to control the integration
domain $V^{\eta}(\underline{w})$ when $\underline{w}$ varies. This
is the purpose of the two propositions which follow, to show that
when the variation $\zeta$ of $\underline{w}$ remains lower than a
given bound, the variations of $F(\underline{w},v)$ (Proposition
1) and that of the integration domain $V^{\eta}$ (Proposition 2)
are negligible compared to the norm of $\zeta$.

\subsubsection{Invariance of the $\psi_p$ functions}

This proposition insures the invariance of the
$\psi_p(\underline{w},v)$ functions for $v$ belonging to
$V^{\eta}(\underline{w})$ and for sufficiently small variations
$\zeta$ of the prototypes $\underline{w}$.

\
\\ \\ \textbf{Proposition 1}: For $\zeta=(\zeta_1,...,\zeta_n) \in
(\mathbb{R}^d)^n$ with
$\underline{w}+\zeta=(w_1+\zeta_1,...,w_n+\zeta_n) \in V^n$, we
denote
$\mid\zeta\mid=\textrm{max}_{p=1,n}\parallel\zeta_p\parallel$, and
$d_{r}^{\zeta}=\parallel(w_r+\zeta_r)-v\parallel$. Thus, we have:
\begin{align}
&\exists\;\nu>0 \;\textrm{such that} \;\forall r,p=1,...,n \;\;
\textrm{for}\; \mid\zeta\mid<\nu: \notag\\
&H((d_{r}^\zeta)^2-(d_{p}^\zeta)^2)=H(d_{r}^2-d_{p}^2), \forall v
\in V^\eta(\underline{w})
\end{align}

\textit{Proof }:

The proof is based on the existence of a bound denoted $\nu$
inside which the invariance of the Heaviside function according to
$\zeta$ is insured. First, we consider the case where the
Heaviside function takes the value 1 and then the case where it is
0:

(i) Considering $H(d_{r}^2\!-\!d_{p}^2)\!=\!1$, we must find a
condition on $\zeta$ for which\\
$(d_{r}^\zeta)^2\!-\!(d_{p}^\zeta)^2\!>\!0$:

For $r\neq p$, we have $d_{r}^2-d_{p}^2 > \eta^2$ and
\begin{align}
(d_{r}^\zeta)^2-(d_{p}^\zeta)^2 &=d_r^2 - d_p^2+2\langle w_r
-v\mid\zeta_r \rangle - 2\langle w_p -v\mid\zeta_p \rangle +
\zeta_r^2-\zeta_p^2
\end{align}
where $\langle.\mid.\rangle$ denotes the scalar product. However,
for the
 scalar products, we have:
\begin{align*}
\langle w_r -v\mid\zeta_r \rangle \geq -2\delta \|\zeta_r\|
\end{align*}
and
\begin{align*}
\langle v -w_p\mid\zeta_p \rangle \geq -2\delta \|\zeta_p\|
\end{align*}
hence
\begin{align*}
(d_{r}^\zeta)^2-(d_{p}^\zeta)^2\geq
&\eta^2-4\delta(\|\zeta_r\|+\|\zeta_p\|)+\zeta_r^2+\zeta_p^2
\end{align*}

Finally, we have:
\begin{align*}
(d_{r}^\zeta)^2-(d_{p}^\zeta)^2\;&\geq\;
(\frac{1}{2}\eta^2-\zeta_p^2-4\delta\parallel\zeta_p\parallel)
+(\frac{1}{2}\eta^2+\zeta_r^2-4\delta\parallel\zeta_r\parallel)\\
&=a_1+a_2
\end{align*}

(ii) Considering $H(d_{r}^2\!-\!d_{p}^2)\!=\!0$, we must find a
condition on $\zeta$ for which\\
$(d_{p}^\zeta)^2\!-\!(d_{r}^\zeta)^2\!>\!0$:

A similar calculation leads to:
\begin{align*}
(d_{p}^\zeta)^2-(d_{r}^\zeta)^2\;&\geq\;
(\frac{1}{2}\eta^2+\zeta_p^2-4\delta\parallel\zeta_p\parallel)
+(\frac{1}{2}\eta^2-\zeta_r^2-4\delta\parallel\zeta_r\parallel)\\
&=b_1+b_2
\end{align*}

The joint study of the polynomials in $\parallel\zeta_p\parallel$
defined by $a_1$ and $b_1$ shows that the conditions $a_1 \geq 0$
et $b_1 \geq 0$ are reached for:

\begin{equation*}
\parallel\zeta_p\parallel\;\leq\;\mu=
(4\delta^2+\frac{\eta^2}{2})^{\frac{1}{2}}-2\delta
\end{equation*}

Moreover,
$a_2\;\geq\;(\frac{1}{2}\eta^2-4\delta\parallel\zeta_r\parallel)$
and
$b_2\;\geq\;(\frac{1}{2}\eta^2-5\delta\parallel\zeta_r\parallel)$

The conditions $a_2\;\geq\;0$ et $b_2\;\geq\;0$ are met for
$\parallel\zeta_r\parallel\;\leq\;\frac{\eta^2}{10\delta}$.

It is enough to take
$\nu=\textrm{min}\{\mu,\frac{\eta^2}{10\delta}\}$.

The proposition 1 means that considering a variation of the norm
of $\underline{w}$ vectors lower than $\nu$, $\psi_p$ functions
remain the same either within the energy function or within the
adaptation rule. As a consequence, the function
$F(\underline{w},v)$ to integrate, which is a combination of
$\psi_p$ functions with continuous functions of $\underline{w}$,
is continuous and differentiable over $V^{\eta}(\underline{w})$
according to $\underline{w}$. The nature of $P$ as being
continuous or not, does not affect this result because $P$ does
not depend on $\underline{w}$.

\subsubsection{Variations of the integration domain}

To study the variations of the energy function, it is necessary to
study the variations of the integration domains.

This proposition insures that the variations of the integration
domains $V^{\eta}(\underline{w})$ and $V\setminus
V^{\eta}(\underline{w})$ remain small with small variations
$\zeta$ of the prototypes. \
\\
\\ \textbf{Proposition 2}: for $\mid \zeta \mid \ll 1$, we have :
\begin{flushleft}
(i) \hspace{15mm} $\mid meas(V \setminus
V^\eta(\underline{w}+\zeta))-
     meas(V \setminus V^\eta(\underline{w}))\mid=\emph{O}(\mid \zeta \mid^2)
     $;\notag\\
(ii) \hspace{14mm} $\mid
meas(V^\eta(\underline{w}+\zeta))-
     meas(V^\eta(\underline{w}))\mid=\emph{O}(\mid \zeta \mid^2) $. \notag
\end{flushleft}

\textit{Proof }: The proof of both equations is obtained by
calculating the measure of the tubular neighborhood
$T_p^{\eta}(\underline{w})$ of the Vorono\"{\i} cells. The
projections of these neighborhoods onto the coordinate axes verify
:

\hspace{15mm}$\mid
meas(T_p^{\eta}(\underline{w}+\zeta)-meas(T_p^{\eta}(\underline{w})
\mid=\emph{O}(\mid \zeta \mid^2)$

In just the same way as in property (\ref{mesureenodeeta}), we can
write:
\begin{align*}
\hspace{15mm}&\mid meas(V \setminus V^\eta(\underline{w}+\zeta))-
meas(V \setminus V^\eta(\underline{w})) \mid \\
&=\sum_{p=1}^n(meas(T_p^{\eta}(\underline{w}+\zeta))
-meas(T_p^{\eta}(\underline{w})))=\emph{O}(\mid \zeta \mid^2)
\end{align*}

validating item (i) of the proposition.

Item (ii) is validated observing that:
\begin{equation*}
meas(V^\eta(\underline{w}))+meas(V \setminus
V^\eta(\underline{w}))=meas(V^\eta(\underline{w}+\zeta))+meas(V
\setminus V^\eta(\underline{w}+\zeta))
\end{equation*}

Hence, for small variations $\zeta$ of $\underline{w}$, the
variations of the integration domains remain negligible compared
to $\zeta$.

\subsubsection{Last step for the proof} \label{sectProofTheorem}

We show that small variations $\zeta$ of $\underline{w}$ ({\it
i.e.} less than the bound $\nu$ determined in Proposition 1) lead
to a small variation of $E_V^\eta$ which breaks down in a linear
application plus other terms of higher order, hence that
$E_V^{\eta}$ is a potential for all $\underline{w}\in V$ and all
$v \in V^{\eta}(\underline{w})$. Then we show that
$E_V(\underline{w})-E_V^\eta(\underline{w})=\emph{O}(\eta),
\;\forall \underline{w} \in V^n\;$ hence that
$\lim_{\eta\rightarrow 0}\parallel
E_V-E^{\eta}_V\parallel_{\infty}=0$ demonstrating that $E_V$ is a
pseudo-potential, and at the same time that $\parallel
E_{V}(\underline{w})-E^{\eta}_{V}(\underline{w})\parallel_{\infty}=\emph{O}\:(\eta)$.

The difference
$E_V^\eta(\underline{w}+\zeta)-E_V^\eta(\underline{w})$ may be
written as:
\begin{align*}
&E_V^\eta(\underline{w}+\zeta)-E_V^\eta(\underline{w})\\
&=[E_V^\eta(\underline{w}+\zeta)-\int_{V^{\eta}(\underline{w})}F(\underline{w}+\zeta,v)dv]\;
+\;[\int_{V^{\eta}(\underline{w})}F(\underline{w}+\zeta,v)dv-E_V^\eta(\underline{w})]\\
&=\hspace{22mm}[part\,1]\hspace{23mm}+\hspace{22mm}[part\,2]
\end{align*}
The function $F(\underline{w},v)$ being bounded on
$V^n\,\times\,V$, Proposition 2 shows that [part\,1] is
$\emph{O}(\mid \zeta \mid^2)$.\\ Proposition 1 leads to:

\begin{align*}
[part\,2]=&-\langle\zeta_j\mid\int_{V^{\eta}(\underline{w})}P(v)\psi_{j}(\underline{w},v)(v-w_{j})dv\rangle\\
&\hspace{25mm}+\frac{\parallel\zeta_j\parallel^2}{2}
\int_{V^{\eta}(\underline{w})}P(v)\psi_{j}(\underline{w},v)(v-w_{j})^2dv
\end{align*}
The first term is of the form $L(\zeta)$, where $L$ is a linear
application and the second term is of higher order. Thus, we can
write:
$E_V^{\eta}(\underline{w}+\zeta)-E_V^{\eta}(\underline{w})=L(\zeta)+O(|\zeta|^2)$,
which means that $E_V^{\eta}(\underline{w})$ is differentiable for
all $\underline{w}\in V$ and $v\in V^{\eta}(\underline{w})$.

Moreover, because the function $F(\underline{w},v)$ is bounded on
$V^n\,\times\,V$, we can write:
\begin{align*}
  E_V(\underline{w})-E_V^\eta(\underline{w})&=\int_{V \setminus V^{\eta}(\underline{w})}F(\underline{w},v)dv \\
  & \leq \sup_{(\underline{w},v) \in V^n\times
  V}{F(\underline{w},v)}.meas(V \setminus V^{\eta})
\end{align*}
whence, with the property (\ref{mesureenodeeta}) :
$E_V(\underline{w})-E_V^\eta(\underline{w})=\emph{O}(\eta)$, for
all $\underline{w}\in V^n$.

The energy function $E_V$ is then a pseudo-potential.

\subsection{A pseudo-potential is not a potential in general}

As far as the neighborhood functions $\psi_p$ are of the form
given in (\ref{functionpsi}), the theorem ensures that the corresponding
energy function is a pseudo-potential over the entire domain $V$,
and at least a potential over $V^{\eta}(\underline{w})$. We also know
 that it exists energy functions 
in this framework ({\it i.e.} pseudo-potentials) which are potential over the entire
domain $V$ for continuous pdf $P$, {\it e.g.} the energy function of the K-means (see section \ref{sectExistingRuleKMEANS}) 
\cite{Pages93}. 
However, it remains to prove the existence of energy functions in this framework
which are not potential over the entire domain $V$, {\it i.e.} the existence 
of pseudo-potentials which are not potentials.

Here we show that the energy function of the K-means does not correspond
to a global potential for a particular discontinuous pdf $P$, hence 
is not a potential in general for all $P$.

In order to simplify the calculi, 
we consider only $n=2$ prototypes $\underline{w}=(w_1,w_2)$ in a
$1$-dimensional space ($d=1$). It is straightforward, though messy, to extend
this result to higher dimensions and greater number of prototypes .

The neighborhood function of the K-means, associated to each
prototype is defined as:
\begin{align*}
&\psi_1(\underline{w},v)=H(d_{2}^2\!-\!d_{1}^2)=H(\parallel w_2\!-\!v
\parallel^2-\parallel w_1\!-\!v\parallel^2)\\
&\psi_2(\underline{w},v)=H(d_{1}^2\!-\!d_{2}^2)=H(\parallel w_1\!-\!v
\parallel^2-\parallel w_2\!-\!v\parallel^2)
\end{align*}

that we shorten $\psi_1(v)$ and $\psi_2(v)$ respectively. We have $\psi_i(v\in V_i)=1$ and $\psi_i(v\not\in V_i)=0$. 

These functions are part of the family given by equation
(\ref{functionpsi}), hence the corresponding energy function $E_V$
is a pseudo-potential and $E_V^\eta$ is a potential. Observing
that $E_V=(E_V-E_V^\eta)+E_V^\eta$, 
we are going to show that
$E_V$ is not a potential by showing that $(E_V-E_V^\eta)$ is not a
potential. The function $(E_V-E_V^\eta)$ is not a potential wrt $\underline{w}$ iff 
the variation of this function wrt some variation $\zeta$ of $\underline{w}$ cannot be written
as $L(\zeta)+O(\zeta^2)$, \textit{i.e.} as a linear form of $\zeta$. 

\begin{figure}
  \center
  \resizebox{5in}{!}{\includegraphics{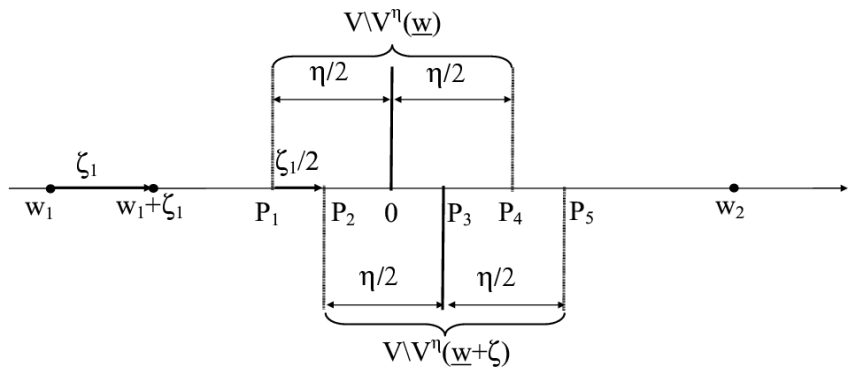}} 
  \caption{\label{figCounterExample}\textbf{Variation of the
  integration domain with $\zeta_1$ }. The point $O$
  is the Vorono\"\i\ boundary between $w_1$ and $w_2$. The point $P_3$
  is the Vorono\"\i\ boundary between
  $w_1+\zeta_1$ and $w_2$.}
  \end{figure}

Let $(w_1+w_2)/2$ be the origin $0$ of the directed line $(w_1w_2)$.
For a small positive variation $\zeta=\zeta_1$ of $w_1$ (see figure
\ref{figCounterExample}), with $0 <
\zeta_1 < \eta$, we have:
\begin{equation}
\label{deltaEnergyKmeans}
\begin{array}{rl}
\Delta(\zeta)=&(E_V\!-\!E_V^\eta)(\underline{w}+\zeta)-(E_V\!-\!E_V^\eta)(\underline{w})\\
=&\displaystyle\int_{V\setminus V^\eta(\underline{w}+\zeta)}F(\underline{w}+\zeta,v)dv\;-\;
\int_{V\setminus V^\eta(\underline{w})}F(\underline{w},v)dv\\
\\
=&\displaystyle\frac{1}{2}\int_{V\setminus
V^{\eta}(\underline{w}\!+\!\zeta)}P(v)\left[\psi_1^{\zeta_1}(v)\delta_1^{\zeta_1}(v)+\psi_2^{\zeta_1}(v)\delta_2(v)\right]dv\\
&-\displaystyle\frac{1}{2}\int_{V\setminus
V^{\eta}(\underline{w})}P(v)\left[\psi_1(v)\delta_1(v)+\psi_2(v)\delta_2(v)\right]dv
\end{array}
\end{equation}
where 
\begin{align*}
&\delta_1(v)=d_1^2=\|w_1\!-\!v\|^2,\\
&\delta_1^{\zeta_1}(v)=\|w_1\!+\!\zeta_1\!-\!v\|^2,\\
&\delta_2(v)=d_2^2=\|w_2\!-\!v\|^2,
\end{align*}
and 
\begin{align*}
&\psi_1^{\zeta_1}(v)=H(\delta_2(v)-\delta_1^{\zeta_1}(v))\\
&\psi_2^{\zeta_1}(v)=H(\delta_1^{\zeta_1}(v)-\delta_2(v)).
\end{align*}

The domains $V\setminus V^{\eta}(\underline{w}+\zeta)$ and
$V\setminus V^{\eta}(\underline{w})$ are defined on the figure
\ref{figCounterExample} and given below:

\begin{align*}
\left\{\begin{array}{lll}
V\setminus V^\eta(\underline{w}+\zeta)&=&[P_2,0]\cup[0,P_3]\cup[P_3,P_4]\cup[P_4,P_5]\\
V\setminus V^\eta(\underline{w})&=&[P_1,P_2]\cup[P_2,0]\cup[0,P_3]\cup[P_3,P_4]
\end{array}
\right.
\end{align*}

where
\begin{align*}
&P_1=-\eta/2\\
&P_2=-\eta/2+\zeta_1/2\\
&P_3=\zeta_1/2\\
&P_4=\eta/2\\
&P_5=\eta/2+\zeta_1/2\\
\end{align*}
 
and $0 < \zeta_1 < \eta$ leads to $P_1<P_2<0<P_3<P_4<P_5$.

Let us consider a particular uniform density $P(v)$ defined as:

\begin{equation}
\begin{array}{ll}
P(v)=&\left\{
\begin{array}{l}
p=\frac{1}{\beta-\lambda} \textrm{ if } v\in [\lambda, \beta] \textrm{ with }  \lambda\in[0,P_4] \textrm{ and } \beta>>\eta \\
0 \textrm{ else}

\end{array}
\right.
\end{array}
\end{equation}

Notice that $P$ is a discontinuous pdf at $\lambda$ and $\beta$. We have also $0<\zeta_1<\eta<<\beta$ hence $P_5<<\beta$.
Then, for such a density and from (\ref{deltaEnergyKmeans}) we get:

\begin{equation}
\label{eqDeltaKmeans}
\displaystyle
\begin{array}{rl}
\displaystyle\Delta(\zeta)=&\displaystyle \int_{P_2}^{P_5}F(\underline{w}+\zeta,v)dv\displaystyle-\int_{P_1}^{P_4}F(\underline{w},v)dv\\
\\
=&\left\{
\begin{array}{l}
\displaystyle\frac{p}{2}\int_{\lambda}^{P_3} (\delta_1^{\zeta_1}(v)-\delta_2(v)) dv + \frac{p}{2} \int_{P_4}^{P_5}\delta_2(v) dv
\textrm{ if } \zeta_1>2\lambda \quad ({\it i.e.}\;\lambda<P_3)\\
\\
\displaystyle\frac{p}{2} \int_{P_4}^{P_5}\delta_2(v) dv \textrm{ if } \zeta_1\leq2\lambda \quad ({\it i.e.}\; \lambda\geq P_3)
\end{array}
\right.
\end{array}
\end{equation}

Developping equation (\ref{eqDeltaKmeans}) leads to:

\begin{equation}
\displaystyle
\begin{array}{rl}
\displaystyle\Delta(\zeta)=&\left\{
\begin{array}{l}
\displaystyle\frac{p}{6}\left[(\lambda\!-\!w_2)^3\!-\!(\lambda\!-\!w_1)^3\!+\!w_2^3\!-\!w_1^3\right]\\
\displaystyle\quad+\frac{p}{4} \left[ 2(w_1\!-\!\lambda)^2\!+\!(w_2\!-\!\frac{\eta}{2})^2\!-\!(w_1^2\!+\!w_2^2)\right] \zeta_1+\!o(\zeta_1^2) \textrm{ if } \zeta_1\!>\!2\lambda \\
\\
\displaystyle\frac{p}{4} (w_2\!-\!\frac{\eta}{2})^2\zeta_1\!+\!o(\zeta_1^2) \textrm{ if } \zeta_1\!\leq \!2\lambda
\end{array}
\right.\\
\\
=&\left\{
\begin{array}{l}
\displaystyle L_1(\zeta_1)+o(\zeta_1^2) \textrm{ if } \zeta_1>2\lambda \\
\\
\displaystyle L_2(\zeta_1)+o(\zeta_1^2) \textrm{ if } \zeta_1\leq2\lambda
\end{array}
\right.

\end{array}
\end{equation}

with $L_1\neq L_2$. Therefore $\Delta(\zeta)$ is not a linear form of $\zeta$
which proves the non differentiability of $(E_V-E_V^\eta)$.
Hence
$(E_V-E_V^\eta)$ is not a potential and so, the energy function $E_V$
is a pseudo-potential but not a potential in general.

\subsection{What is important about this result}

{\bf Consequence 1}: The family of pseudo-potential functions 
includes potential functions as a special case and it exists pseudo-potential functions which are not potentials.

{\bf Consequence 2}: The previous example shows that a necessary condition for the energy function of the K-means to be a potential is that given 
any number $n$, position $\underline{w}$ and dimension $d$ of the prototypes, the boundary of Vorono\"\i\ cells never crosses any discontinuity of the pdf $P$ 
whatever the value of the variation $\zeta$. A sufficient condition for this to hold is $P$ being continuous.

{\bf Consequence 3}: The energy function of the K-means is not a potential at least for some discontinuous pdf $P$. This complements 
the result of Pag\`es \cite{Pages93} stating this energy function is a potential for continuous $P$.
Moreover, the algorithms presented in section \ref{sectExistingRule}, because they reduce to the K-means for specific values 
of their parameters, also share this property that prevent them from being potentials in general for all $P$ and all setting of their
parameters.
In particular, this result holds for the Neural-Gas \cite{Martinetz93} despite the claim of its authors: the Neural-Gas is not a
global potential at least for discontinuous $P$ and width $\sigma$ of the neighborhood function set to $0$. 
This casts some doubt on the validity of their proof which do not specify any restriction on $P$ and $\sigma$. 
As a consequence, the convergence of the associated adaptation rule in general still to be proved.

\subsection{Consequence of the theorem concerning the convergence}
\label{conseqconverg}
 The consequence of the theorem is promising
 concerning the eventual convergence of the adaptation rules associated to pseudo-potentials 
 toward a local minimum.
 Indeed, from a mathematical point of view, talking about
 "derivatives" of the energy function $E_V(\underline{w})$ onto $V$ according to some $w_p$
 does not make any sense because of the discontinuities of this function onto
 the Vorono\"\i\ boundaries. The only possibility is to measure the variations
 of this function according to a small movement of the prototypes.
 We have already shown that the volume of the tubular neighborhood 
 of the Vorono\"\i\ boundaries is
 in $O(\eta)$ (Equation (\ref{mesureenodeeta})) so is bounded. Now this theorem
 shows that the variations of the energy
 function according to a bounded movement $\zeta$ of the prototypes, are also bounded. 
 
 Indeed, the theorem allows to write that $E_V(\underline{w}\!+\!\zeta)\!-\!E_V^{\eta} (\underline{w}\!+\!\zeta)\!=\!O(\eta)$ 
 and $E_V(\underline{w})\!-\!E_V^{\eta} (\underline{w})\!=\!O(\eta)$ so 
 $[E_V(\underline{w}\!+\!\zeta)\!-\!E_V(\underline{w})]\!-\![E_V^{\eta} (\underline{w}\!+\!\zeta)\!-\!E_V^{\eta}
 (\underline{w})]\!=\!O(\eta)$.
 And $E_V^{\eta}$ being a potential, then $E_V^{\eta} (\underline{w}\!+\!\zeta)\!-\!E_V^{\eta} (\underline{w})$ 
 is bounded as a linear form of $\zeta$ which is bounded. Therefore
 $E_V(\underline{w}\!+\!\zeta)\!-\!E_V(\underline{w})\!=\!\Delta_V(\zeta)$ is also bounded
 although $E_V$ is not a potential on $V$.
 As a consequence, the effects of the variation $\Delta_V(\zeta)$ of the energy function $E_V$ 
 according to $\zeta$, on the dynamic of the prototypes
 remains negligible on average even for some data falling onto the Vorono\"\i\ boundaries.
 In other words, the existence of a pseudo-potential for lack of a potential would be 
 sufficient to ensure the convergence of the associated adaptation rule, 
 although a rigorous proof is still to be carried out. The work of Bottou \cite{Bottou91}
 gives also insights in this direction but following a different way.

\section{Consequence for existing rules}
\label{sectExistingRule}

In this section, we show that the neighborhood function of a large
number of  algorithms can be written in the form of the equation
(\ref{functionpsi}), \textit{i.e.} as a combination of Heaviside
step functions of a difference of squared distances $d_{l}$. This
demonstrates that the corresponding adaptation rule is associated
to an energy function which is not necessarily a potential but at
least a pseudo-potential.

\subsection{K-means vector quantizer}
\label{sectExistingRuleKMEANS}

The K-means vector quantizer \cite{MacQueen67} is the iterative
version of the Linde-Buzzo-Gray batch learning technique for
vector quantization \cite{Linde80}. It consists in presenting one
datum $v$ at a time, then selecting the closest prototype
$w_{p^*}$ to it and moving it toward $v$. The corresponding
neighborhood function can be written as:

\begin{equation}
\label{eqKmeans}
\begin{array}{rl}
\psi_p^{[\textrm{K-means}]}(\underline{w},v)=&A_p(\underline{w},v)
=\displaystyle K_p(\underline{w},v)\prod_{l=1}^{n}(K_l(\underline{w},v)(H(l-p)-1)+1)\\
=&\left\{\begin{array}{l}
1 \textrm{  if  } v\in V_p \textrm{  and  } p=\min(\{i\in (1,\dots,n)|K_i(\underline{w},v)=1\})\\
0 \textrm{  else}
\end{array}
\right.

\end{array}
\end{equation}

where the function $K_p$ is an indicator function of the
Vorono\"\i\ cell $V_p$ of $w_p$, defined as:

\begin{equation}\label{K_p}
\begin{array}{rl}
 K_p(\underline{w},v)=
&\displaystyle\prod_{k=1}^{n}H(d_{k}^2-d_{p}^2) \\
=&\left\{\begin{array}{l}
1 \textrm{  if  } v\in V_p\\
0 \textrm{  else}
\end{array}
\right.

\end{array}
\end{equation}

The function $A_p$ performs an additional sort over the index of
the closest prototypes (the "winners") for which $K_p$ is
equal to $1$, \textit{i.e.} all the prototypes which are
the closest to $v$. This is the algebraic writing of the
algorithms which choose only one prototype among all the
closest one in case of equality. Here, the choice is carried out
according to the lowest index, it could be the highest one, or a
random choice among the indices of all the winners. In case where
all the winners are moved, then
$\psi_p^{[\textrm{K-means}]}(\underline{w},v)=K_p(\underline{w},v)$
should be considered.

The K-means algorithm corresponds to a Hard Competitive Learning
technique \cite{Ahalt90}, where only the closest prototype
to the datum is adapted at a time. To escape from local optima of
the energy function, it has been improved by defining a
neighborhood function which enables the winner to be adapted and
also some of its neighbors. All the following algorithms belong to
that class of Soft-Competitive Learning techniques \cite{Ahalt90},
and each one defines its particular neighborhood function.

\subsection{Self-Organizing Maps and other graph-based neighborhoods}

The Self-Organizing Map (SOM) proposed by Kohonen \cite{Kohonen82}
defines a set of connections between the prototypes, which
corresponds to a graph G with a particular topology (\textit{e.g.}
a regular 2-dimensional grid). The winner being determined
according to the datum $v$, the neighborhood function consists in
weighting the adaptation step of the prototypes according
to their closeness to the winner on the graph G.

The corresponding neighborhood function may be written as:

\begin{equation}
\psi_p^{[\textrm{SOM}]}(\underline{w},v)=\sum_{q=1}^{n}A_q(\underline{w},v)h_{\sigma}(D_{qp}(G))
\end{equation}

where $h_{\sigma}$ is a non-increasing positive function with a
tunable width $\sigma$ (\textit{e.g.}
$h_{\sigma}(u)=e^{-\frac{u}{\sigma}}$) and $D_{ab}(G)$ is the
distance between $w_a$ and $w_b$ in terms of the lowest number of
edges separating them within the graph G.

Several other algorithms essentially differ from SOM by the fact
they use a graph whose topology is not defined \textit{a priori}
but thanks to the data and the prototypes positions in the
data space. This is the case in the Growing Neural-Gas (GNG) of
\cite{Fritzke95a}, where G is the Induced Delaunay Triangulation
(IDT) \cite{Martinetz94}, in \cite{Kangas90} with Minimum Spanning
Trees (MST), in \cite{Mou94} with Gabriel Graphs, in the Growing
Cell Structure (GCS) of \cite{Fritzke94} with a set of simplices
with fixed dimension, and in the growing versions of SOM (GSOM) of
\cite{Fritzke95b} and \cite{Villmann97} with an adaptive grid
structure. As far as $n$ remains constant and the graph G remains
the same, the neighborhood function of all these models is identical
to the one of the SOM written above, and belongs to the framework we
consider in this paper.

\subsection{Neural-Gas}

In the Neural-Gas \cite{Martinetz93}, the prototypes are
ranked in increasing order of their distance to the datum $v$.
This rank is used to weight the adaptation rule of the prototypes. 
Martinetz et. al. give the corresponding neighborhood
function :

\begin{equation}
\psi_p^{[\textrm{Neural-Gas}]}(\underline{w},v)=h_{\sigma}(k_p(\underline{w},v))\textrm{
with  }k_p(\underline{w},v)=\sum_{q=1}^n
\Upsilon(d_{p}^2-d_{q}^2) 
\end{equation}
where $\Upsilon(u)=1-H(-u),\; \forall u$. The function $k_p$ is
the rank of the prototype $w_p$ such that
 $k_p(\underline{w},v)=\!j\!-\!1$ iff $p$ is the $j^{\textrm{\small th}}$ closest vector to $v$
 (several prototypes may have the same rank). Note that the Neural-Gas could be included
 into the previous family of adaptive graph-based neighborhoods
considering G as the graph which connects the
$n$-nearest-neighbors of $v$ among $\underline{w}$, in a chain
where the $i^{th}$ nearest neighbor is connected to the
$(i-1)^{th}$ ($\forall i>1$) and the $(i+1)^{th}$ ($\forall i<n$).

\subsection{Recruiting rules}

One of us proposed the ``Recruiting'' Neural-Gas \cite{Aupetit00} as a
way to cope with function approximation tasks using vector
quantizers. A recruiting factor is added to the Neural-Gas
adaptation rule. Such a factor is associated to each prototype
 and the winner imposes its own on the others. This tends to
gather the prototypes around the one which has the highest
recruiting factor. Then setting this factor proportional to the
local output error approximating a function, enables more
prototypes to be grouped together in areas of the input
space where the corresponding output function is more difficult to
approximate. This tends to decrease the global approximation
error.

The corresponding neighborhood function may be written as:

\begin{equation}
\psi_p^{[\textrm{RecruitingNG}]}(\underline{w},v)=h_{\sigma}(k_p(\underline{w},v))\sum_{q=1}^n A_q(\underline{w},v)\epsilon_q
\end{equation}

where $\forall q,\epsilon_q \in[0,1]$.
Taking $\epsilon_q=\epsilon_p=1, \; \forall p,q$ leads to the
usual Neural-Gas.

 G\"oppert and Rosenstiel \cite{Goppert96} proposed a
similar approach with a SOM for which each prototype
defines its own neighborhood's width $\sigma_q$ tuned according to
the local output approximation error. The corresponding
neighborhood function may be written as:

\begin{equation}
\psi_p^{[\textrm{RecruitingSOM}]}(\underline{w},v)=\sum_{q=1}^{n}A_q(\underline{w},v)h_{\sigma_q}(D_{qp}(G))
\end{equation}

where $\forall q, \sigma_q\in[0,1]$. 
Taking $\sigma_q=\sigma_p=\sigma, \; \forall p,q$ leads to the
usual SOM.

In both approaches, as far as $\epsilon_q$ and $\sigma_q$ remain
independent of $v$ and $\underline{w}$, the corresponding
neighborhood function belongs to the framework we consider in this
paper.


\subsection{Concerning the algorithms with adaptive structures}

We have shown that many vector quantization algorithms belong to
our framework. However, considering dynamic approaches such as the
algorithms which adapt either the number $n$ of prototypes (GCS,
GNG, GSOM), the graph of their neighborhood structure (GNG,
GSOM), or the recruiting factor (RecruitingNG, RecruitingSOM),
according to either the number of iterations, the position of the
prototypes or the output approximation error, it is still
difficult to define a framework taking into account these
structural changes. That is why we considered these dynamic
parameters to be fixed in such cases.

\subsection{About some algorithms which do not belong to the present framework}

We shall notice that the modified Self-Organizing Map proposed by
Heskes and Kappen \cite{Heskes93,Heskes99} does not belong to the
present framework. Indeed the Heaviside step functions involved in
the corresponding $\psi_p$ neighborhood functions are not applied
to a pair of square distances $d_{l}$ directly, but to a sum over
$\underline{w}$ of weighted square distances $d_{l}$. This
prevents $\psi_p$ from belonging to the family we consider in
equation (\ref{functionpsi}). However it seems possible to
enlarge our framework in order to encompass the neighborhood
function they proposed.

The $\gamma$-Observable Neighborhood has been proposed by one of
us \cite{Aupetit02} as a neighborhood that decreases the
number of iterations needed for the adaptation rule to converge
toward an optimum of the energy function. The corresponding
neighborhood function does not belong to the present framework.
However, we have already defined an extension of this framework
which encompasses this adaptation rule and thus which allows to
demonstrate that the energy function associated to the
$\gamma$-Observable Neighbors is also a pseudo-potential. This work has not been
 published yet.

\section{Conclusion}

In vector quantization, we propose a framework which ensures the
existence of a family of potential functions ({\it i.e.} differentiable functions)
which converges
uniformly to the energy function that we call in such a case a
"pseudo-potential". We demonstrate that a pseudo-potential is not
necessarily differentiable everywhere, hence it is not always a
potential. As a consequence, the corresponding adaptation rule
does not necessarily perform a stochastic gradient descent along
this energy function.

We also show how a large number of existing vector quantization
algorithms belong to this framework, hence
even if they are not associated to potentials, they are at least associated to pseudo-potentials. 
This framework allows to study at once the convergence of all these algorithms.
At that point, although the pseudo-potentials are not necessarily potentials, 
a consequence of the theorem shows that the variations of the pseudo-potentials
on the boundaries of the Vorono\"\i\ cells remain bounded, so they have 
a negligible effect on the dynamic of the prototypes on average. This is 
a promising preliminary result about the convergence of the corresponding adaptation rules.

If the convergence of the adaptation rules associated to pseudo-potentials were demonstrated 
then the present framework would constitute an {\it a posteriori}
justification of a large family of adaptation rules considered up
to now as heuristic. Moreover, this framework makes possible the
design of new adaptation rules respecting the hypotheses which ensure
the existence of the corresponding pseudo-potential. 

 The results of this paper suggest two avenues for future
 research:
 \begin{itemize}
 \item investigating the convergence properties of the
adaptation rules associated to pseudo-potentials in general.
\item extending this framework to a wider class of
neighborhood functions. 
\end{itemize}

By introducing pseudo-potentials, we add a new concrete framework
on the wasteland of non-potentials. Within this framework, the consequence of the theorem makes us hopeful to
build new theorems which could insure at once the convergence 
with respect to a specific norm, of a large
number of existing vector quantization algorithms which are not 
associated to potentials but at least to pseudo-potentials.

\bibliographystyle{apalike}

\newpage

\bibliography{submitArxiv04_2006}

\newpage

\end{document}